\documentclass{article}
\usepackage{ijcai15}

\usepackage{graphicx}
\usepackage{algorithm}
\usepackage{algorithmic}
\usepackage{amsthm}
\usepackage{amssymb}
\usepackage{multicol}
\usepackage{amsmath}

\usepackage{bbm,amssymb,amsmath,amsfonts,graphicx,fullpage,ifthen,twoopt,algorithm,algorithmic,color}

\usepackage{times}

\title{Context Attentive Bandits:\\
	 Contextual Bandit with Restricted Context}
\author{Djallel Bouneffouf$^1$, Irina Rish$^2$, Guillermo A. Cecchi$^3$, Rapha\"{e}l F\'{e}raud$^4$\\
$^{1,2,3}$IBM Thomas J. Watson Research Center, Yorktown Heights, NY USA\\
$^4$Orange Labs, 2 av. Pierre Marzin, 22300 Lannion (France)\\
\{dbouneffouf, Irish, gcecchi \}@us.ibm.com \\
Raphael.Feraud@orange.com
}

\begin{document}

\maketitle

\begin{abstract}
We consider a novel formulation of the multi-armed bandit model, which we call the {\em contextual bandit with restricted context}, where only a limited number of features can be accessed by the learner at every iteration. This novel formulation is motivated by different on-line problems arising in clinical trials, recommender systems and  attention modeling. Herein, we adapt the standard multi-armed bandit algorithm known as Thompson Sampling to take advantage of our restricted context setting, and propose two novel algorithms, called the {\em Thompson Sampling with Restricted Context (TSRC)} and the {\em Windows Thompson Sampling with Restricted Context (WTSRC)}, for handling stationary and nonstationary environments, respectively. 
Our  empirical results demonstrate advantages of the proposed approaches  on several real-life datasets.
 
\end{abstract}

\section{Introduction}
In sequential decision problems encountered in various applications, from clinical trials \cite{villar2015multi} and recommender systems  \cite{MaryGP15} to visual attention models \cite{mnih2014recurrent}, a decision-making algorithm must choose among several actions at each time point. The actions are typically associated with a side information, or a context (e.g., a user's profile), and the reward feedback is limited to the chosen option. For example, in image processing with attention models \cite{mnih2014recurrent},  the context is an input image,  the action is classifying the image into one of the given categories,  and the reward is 1 if classification is correct and 0 otherwise. A different example involves clinical trials \cite{villar2015multi}, where  the context is the patient's medical record (e.g. health condition, family history, etc.), the actions correspond to the treatment options being compared, and the reward represents the outcome  of the proposed treatment (e.g., success or failure). In this setting, we are looking for a good trade-off between the exploration (e.g., of the new drug) and the exploitation (of the known drug).

This inherent exploration vs. exploitation trade-off exists in many sequential decision problems, and is traditionally formulated  as the {\em multi-armed bandit (MAB)} problem, stated as follows: given $K$ ``arms'', or  possible actions, each associated with a fixed but unknown reward probability distribution ~\cite {LR85,UCB},  an agent selects an arm to play at each iteration, and receives a reward,  drawn according to the selected arm's distribution, independently from the previous actions. A particularly useful version of MAB is the {\em contextual multi-arm bandit (CMAB)}, or simply {\em contextual bandit} problem, where at  each iteration,  before choosing an arm, the agent observes an $N$-dimensional {\em context}, or {\em feature vector}.
The agent uses this context, along with the rewards of the arms played in the past, to choose which arm to play in the current iteration. Over time, the agent's aim is to collect enough information about the relationship between the context vectors and rewards, so that it can predict the next best arm to play by looking at the corresponding contexts \cite{langford2008epoch,AgrawalG13}.

We introduce here a novel and practically important special case of the contextual bandit problem, called the {\em contextual bandit with restricted context (CBRC)}, where observing the full feature vector at each iteration is too expensive or impossible for some reasons,  and the player can only request to observe a limited number of those  features; the upper bound (budget) on the feature subset is fixed for all iteration, but within the budget, the player can choose any feature subset of the given size.  The  problem is to select the best feature subset so that the overall reward is maximized, which involves  exploring both the feature space as well as the arms space.

For instance, in \cite{TekinAS15}, the analysis of clinical trials shows that a doctor can ask a patient only a limited number of question before deciding on drug prescription. In the visual attention models involved in visual pattern recognition \cite{mnih2014recurrent}, a retina-like representation is used, where at each moment only a small region of an image is observed at high resolution, and image classification is performed based on such partial information about the image. Furthermore, in recommender system setting \cite{hu2011nextone}, recommending an advertisement depends on user's profile, but usually only a limited aspects of such profile are available.
The above examples can be modeled within the proposed framework, assuming that only a limited number of features from the full context can be selected and observed  before choosing an arm (action) at each iteration.

Overall, the main contributions of this paper include (1)  a new formulation of a bandit problem with restricted context, motivated by practical applications with a limited budget on information access, (2)  two new algorithms, both for stationary and non-stationary settings of the restricted-context contextual bandit problem,
 and (3) empirical evaluation demonstrating advantages of the proposed methods over a range of datasets and parameter settings.

This paper is organized as follows. Section \ref{sec:related} reviews related works. Section \ref{background} introduces some background concepts. Section \ref{sec:statement} introduces the contextual bandit model with restricted context,  and the proposed algorithms for both stationary and non-stationary environments. Experimental evaluation on several datasets, for varying parameter settings, is presented in Section \ref{sec:experimental}. Finally, the last section concludes the paper and points out possible directions for future works.

\section{Related Work}
\label{sec:related}
The multi-armed bandit problem has been extensively studied. Different solutions have been proposed using a stochastic formulation ~\cite {LR85,UCB,BouneffoufF16} and a Bayesian formulation ~\cite {AgrawalG12}; however, these approaches did not take into account the context.

In LINUCB ~\cite{Li2010}, Neural Bandit \cite{AllesiardoFB14} and in Contextual Thompson Sampling (CTS)~\cite{AgrawalG13}, a linear dependency is assumed between the expected reward of an action and its context; the representation space is modeled using a set of linear predictors.
However, the context is assumed to be fully observable, unlike in this paper.

Motivated by dimensionality reduction task,   \cite{YadkoriPS12} studied a sparse variant of stochastic linear bandits, where only a relatively small (unknown) subset of features is relevant to a multivariate function optimization.
Similarly, \cite{CarpentierM12} also considered the high-dimensional stochastic linear bandits with sparsity, 
combining the ideas from compressed sensing and bandit theory.
In \cite{bastani2015online}, the problem is formulated as a MAB with high-dimensional covariates, and a new efficient bandit algorithm based on the LASSO estimator is presented. 
Still, the above work, unlike ours, assumes  full observability of the context variables.

In classical online learning (non-bandit) setting, where the actual label of a mislabeled sample is revealed to the classifier (unlike 0 reward for any wrong classification decision in the bandit setting),  the authors of~\cite{wang2014online} investigate the problem of Online Feature Selection, where the aim is to make  accurate predictions using only a small number of active features. 

Finally, \cite{durand2014thompson} tackles the online feature selection problem by addressing the combinatorial optimization problem in the stochastic bandit setting with bandit feedback, utilizing the Thompson Sampling algorithm. Note that {\em none of the previous approaches addresses the problem of  context restriction (variable selection) in the contextual bandit setting}, which is the main focus of this work. 

\section{Background}
\label{background}
This section  introduces some background concepts  our approach builds upon, such as contextual bandit, combinatorial bandit, and Thompson Sampling.

\noindent{\bf The contextual bandit problem.}
Following \cite{langford2008epoch}, this problem is defined as follows.
At each time point (iteration) $t \in \{1,...,T\}$, a player is presented with a {\em context} ({\em feature vector}) $\textbf{c}(t) \in \mathbf{R}^N$
  before choosing an arm $k  \in A = \{ 1,...,K\} $.
We will denote by
  $C=\{C_1,...,C_N\}$  the set of features (variables) defining the context.
Let ${\bf r} = (r_{1}(t),...,$ $r_{K}(t))$ denote  a reward vector, where $r_k(t) \in [0,1]$ is a reward at time $t$  associated with the arm $k\in A$.
Herein, we will primarily focus on the Bernoulli bandit with binary reward, i.e. $r_k(t) \in \{0,1\}$.
Let $\pi: C \rightarrow A$ denote a policy.  Also, $D_{c,r}$ denotes a joint distribution  $({\bf c},{\bf r})$.
We will assume that the expected reward is a  linear function of the context, i.e.
$E[r_k(t)|\textbf{c}(t)] $ $= \mu_k^T \textbf{c}(t)$,
where $\mu_k$ is an unknown weight vector (to be learned from the data) associated with the arm $k$. \\

\noindent{\bf Thompson Sampling (TS).}
The TS \cite{thompson1933likelihood}, also known as
Basyesian posterior sampling, is a classical approach to multi-arm bandit problem, where the  reward $r_{k}(t)$ for choosing an arm $k$ at time $t$ is assumed to follow a distribution $Pr(r_{t}|\tilde{\mu})$ with the  parameter $\tilde{\mu}$. Given a prior $Pr(\tilde{\mu})$ on these parameters, their posterior distribution   is given by the Bayes rule, $Pr(\tilde{\mu}|r_{t}) \propto Pr(r_{t}|\tilde{\mu}) Pr(\tilde{\mu})$. A particular case of the Thomson Sampling approach assumes a  Bernoulli bandit problem, with rewards being 0 or 1, and the parameters following the Beta prior.
TS initially assumes arm $k$ to have prior $Beta(1, 1)$ on $\mu_k$ (the probability of success). At time $t$, having observed $S_k(t)$ successes (reward = 1) and $F_k(t)$ failures (reward = 0), the algorithm updates the distribution on $\mu_k$ as $Beta(S_k(t), F_k(t))$. The algorithm then generates independent samples $\theta_k(t)$ from these posterior distributions of the $\mu_k$, and selects the arm with the largest sample value. For more details,  see, for example, \cite{AgrawalG12}. \\


\noindent{\bf Combinatorial Bandit.}
Our feature subset selection aproach will build upon the  {\em combinatorial bandit} problem \cite{durand2014thompson},  specified as follows:
Each arm $k \in \{1,...,K\}$ is associated with the corresponding variable $x_{k}(t)\in R$,  which indicates the reward obtained when choosing the $k$-th arm at time $t$, for $t>1$. Let us consider a constrained set of arm subsets $S \subseteq  P(K)$, where $P(K)$ is the power set of $K$, associated with a set of variables $\{r_{M}(t)\}$, for all $M \in S$ and $t > 1$.
Variable $r_{M}(t)\in R$ indicates the reward associated with selecting a subset of arms $M$ at time $t$, where $r_{M}(t)=h(x_{k}(t))$, $k\in M$, for some function  $h(\cdot)$.
The combinatorial bandit setting can be viewed as a game where a player sequentially selects	subsets in $S$ and observes rewards corresponding to the played subsets. Here we will define the  reward function $h(\cdot)$ used to compute $r_{M}(t)$ as a sum of the outcomes of the arms in $M$, i.e.   $r_{M}(t)=\sum_{k\in M} x_{k}(t)$, although more sophisticated nonlinear rewards are also possible. The objective of the combinatorial bandit problem  is to maximize the reward over time. We consider here a stochastic model, where $x_i(t)$ observed for an arm $k$ are random variables independent
and distributed according to some unknown distribution with unknown expectation $\mu^k$. The outcomes
distribution can be different for each arm. The global rewards $r_{M}(t)$ are also random variables independent and  distributed according to some unknown distribution with unknown expectation $\mu^M$.



\section{Problem Setting}
\label{sec:statement}
In this section, we  define a new type of a bandit problem, the {\em contextual bandit with restricted context (CBRC)};
  the combinatorial task of feature subset (i.e., restricted context) selection
  as treated as a  combinatorial bandit problem \cite{durand2014thompson}, and our approach will be based on the  Thompson Sampling \cite{AgrawalG12}.

\subsection{Contextual Bandit with Restricted Context in Stationary Environment (CBRC)}


Let $\bf c(t) \in \mathbf{R}^N$ denote a  value assignments to the vector of random variables, or features, $(C_1,...,C_N)$ at time $t$,  and let  $C=\{1,...,N\}$ be the set of their indexes. Furthermore, let $\textbf{c}^d(t)$ denote a sparse vector of assignments to only $d \leq N$  out of $N$ features, with indexes from a subset $ C^d \subseteq C$, $|C^d|=d$, and with zeros assigned to all features with indexes outside of $C^d$. 

Formally, we denote the set of all such vectors as  $\mathbf{R}^N_{C^d} = $ $\{\textbf{c}^d(t) \in \mathbf{R}^N |$ $ c^d_i=0~~ \forall i \notin C^d\}$.
In the future, we will always use $C^d$ to denote a feature subset of size $d$, and by $\textbf{c}^d$ the corresponding sparse vector. We will consider a set 
$\Pi^d = \cup_{C^d \subseteq C} \{\pi: \mathbf{R}^N \rightarrow A,~ \pi(\bf c) = \hat{\pi}(s(\bf c))\}$ of compound-function policies, where  the function   $s: \mathbf{R}^N \rightarrow \mathbf{R}^N_{C^d}$ maps each $\bf c(t)$ to  $\textbf{c}^d(t)$, for a given subset $C^d$,  and the function $\hat{\pi}: \mathbf{R}^N_{C^d} \rightarrow A$ maps $\textbf{c}^d(t)$ to an action in $A$.

As mentioned before, in our setting, the rewards are  binary  $r_k(t) \in \{0,1\}$.
The objective of a contextual bandit algorithm would be  learn a hypothesis $\pi$ over $T$ iterations maximizing the total reward. 

\begin{algorithm}[H]
	\caption{ The CBRC Problem Setting}
	\label{alg:CBP}
	\begin{algorithmic}[1]
		\STATE {\bfseries }\textbf{Repeat}
		\STATE {\bfseries } $(c(t),r(t))$ is drawn according to distribution $D_{c,r}$
		\STATE {\bfseries } The player chooses a subset $C^d \subseteq C$
\STATE{\bfseries} The values of $\textbf{c}^d(t)$ of features in $C^d$ are revealed
\STATE{\bfseries} The player chooses an arm $k(t) = \hat{\pi}(\mathbf{c}^d(t))$
				\STATE {\bfseries } The reward $r_{k(t)}$ is revealed
		\STATE {\bfseries } The player updates its policy $\pi$
			\STATE {\bfseries } $t=t+1$
		\STATE {\bfseries }\textbf{Until} t=T
	\end{algorithmic}
\end{algorithm}

\subsubsection{Thompson Sampling with Restricted Context}
\label{subsec:UCB}

\begin{algorithm}[h]
 \caption{Thompson Sampling with Restricted Context (TSRC)}
\label{alg:TSRC}
\begin{algorithmic}[1]
 \STATE {\bfseries }\textbf{Require:} The size $d$ of the feature subset, the initial values $S_i(0)$ and $F_i(0)$ of the Beta distribution parameters. 

 \STATE {\bfseries }\textbf{Initialize:} $\forall k \in \{1,...,K\}, B_k=I_d$,
 $\hat{\mu_k}= 0_d, g_k = 0_d$, and $\forall i \in \{1,...,N\}$, $n_i(0)=0$, $r^f_i(0)=0$.

 \STATE {\bfseries }\textbf{Foreach} $t = 1, 2, . . . ,T$ \textbf{do}

 \STATE {\bfseries }\quad\textbf{Foreach} context feature $i= 1,...,N$ \textbf{do}
 \STATE {\bfseries }\quad\quad $S_i(t)=S_i(0)+r^f_i(t-1)$
 \STATE {\bfseries } \quad\quad$F_i(t)=F_i(0)+n_i(t-1)-r^f_i(t-1)$
 \STATE {\bfseries } \quad\quad Sample $\theta_i$ from $ Beta(S_i(t), F_i(t))$ distribution
\STATE {\bfseries } \quad\textbf{End do}
 \STATE {\bfseries }\quad Select $C^d(t)= argmax_{C^d \subseteq C} \sum_{i \in C^d } \theta_i$
 \STATE  {\bfseries }\quad
 Obtain sparse vector $\textbf{c}^d(t)$ of feature assignments in $C^d$, where
 $c_i=0~~ \forall i \notin C^d$
 \STATE {\bfseries }\quad\textbf{Foreach} arm $k= 1,...,K$ \textbf{do}
 \STATE {\bfseries }\quad \quad Sample $\tilde{\mu_k}$ from $N(\hat{\mu_k}, v^2 B_k^{-1})$ distribution.
 \STATE {\bfseries } \quad\textbf{End do}
 \STATE {\bfseries }\quad Select arm $k(t)= \underset{k\subset \{1,...,K\} }{argmax} \ \textbf{c}^d(t)^\top \tilde{\mu_k} $
 \STATE {\bfseries }\quad Observe $r_{k}(t)$
\STATE {\bfseries }\quad\textbf{if} $r_{k}(t)=1$ \textbf{then}
\STATE \quad\quad$B_k= B_{k}+ \textbf{c}^d(t)\textbf{c}^d(t)^{T} $
\STATE \quad\quad $g_k = g_k + \textbf{c}^d(t)r_{k}(t)$
\STATE \quad\quad$\hat{\mu_k} = B_k^{-1} g_k$
\STATE \quad\quad$\forall i \in C^d(t)$, $n_i(t)=n_i(t-1)+1$ and $r^f_i(t)=r^f_i(t-1)+1$.
\STATE {\bfseries }\quad\textbf{End if}
 \STATE {\bfseries }\textbf{End do}
 \end{algorithmic}
\end{algorithm}

We now propose a method for solving the  stationary CBRC problem, called   {\em Thompson Sampling with Restricted Context (TSRC)}, and summarize it  in Algorithm \ref{alg:TSRC} (see section \ref{background} for background on Thompson Sampling); here the combinatorial task of selecting the best subset of features is approached as a combinatorial bandit problem, following the approach of \cite{durand2014thompson}, and the subsequent  decision-making (action selection) task as a contextual bandit problem solved by Thompson Sampling \cite{AgrawalG13}, respectively.

Let $n_i(t)$ be the number of times the $i$-th feature has been selected so far, let  $r^f_i(t)$ be the cumulative reward associated with the feature $i$, and let $r_k(t)$ be the reward associated with  the arm $k$ at time $t$.

The algorithm takes as an input the desired number of features $d$, as well as the initial values of the Beta distribution parameters in TS.
At each iteration $t$, we update the values of those parameters, $S_i$ and $F_i$ (steps 5 and 6), to represent the current total number of successes and failures, respectively, and then sample the ''probability of success'' parameter $\theta_i$
from the corresponding $Beta$ distribution, separately for each feature $i$ to estimate $\mu^i$, which is the mean reward conditioned to the use of the variable $i: \mu^i = \frac{1}{K} \sum_k E[ r_k . 1\{i \in C^d \}]$ (step 7).
We select the best subset of features, $C^d \subseteq C$, such that $C^d=\arg \max_ {C^d \in C} \sum_{i \in C^d} \theta^i$ (step 9). So the goal of the combinatorial bandit in TSRC algorithm is to maximize:  $E[r_{C^d}] = \sum_{i \in C^d} \mu^i$; note that implementing this step does not actually require combinatorial search\footnote{Since the individual rewards $\theta_i$ are non-negative (recall that they follow Beta-distribution), we can simply select the set $C^d$ of $d$ arms with the highest
individual rewards $\theta_i(t)$.}.

Once a subset of features is selected using the combinatorial bandit approach, we switch to the contextual bandit setting 
in steps 10-13, to choose an arm based on the context consisting now of a subset of features.

We will assume that  the expected reward is a linear function of a restricted context, $E[r_k(t)|\textbf{c}^d(t)] $ $= \mu_k^T \textbf{c}^d(t)$; note that this assumption is different from the linear reward assumption of \cite{AgrawalG13} where single parameter $\mu$ was considered for all arms, there was no restricted context, and for each arm, a separate context vector $c_k(t)$ was assumed. 

Besides this difference, we will follow the approach of \cite{AgrawalG13} for the contextual Thompson Sampling. We assume
that  reward $r_{i}(t)$ for choosing arm $i$ at time $t$   follows a parametric likelihood function $Pr(r(t)|\tilde{\mu_k})$, and that
the posterior distribution at time $t + 1$, $Pr(\tilde{\mu}|r(t)) \propto Pr(r(t)|\tilde{\mu}) Pr(\tilde{\mu})$ is given by a multivariate Gaussian distribution $\mathcal{N}(\hat{\mu_k}(t+1)$, $v^2 B_k(t + 1)^{-1})$, where
$B_k(t)= I_d + \sum^{t-1}_{\tau=1} c(\tau) c(\tau)^\top$
with $d$ the size of the context vectors $c$, $v= R \sqrt{\frac{24}{\epsilon} d ln(\frac{1}{\gamma})}$ with $R>0$,  $\epsilon \in ]0,1]$, $\gamma \in ]0,1]$ constants, and $\hat{\mu}=B(t)^{-1} (\sum^{t-1}_{\tau=1} c(\tau) c(\tau))$.

At each time point $t$, and for each arm, we sample
 a $d$-dimensional $\tilde{\mu_k}$ from
$\mathcal{N}(\hat{\mu_k}(t)$, $ v^2B_k(t)^{-1})$,  and choose an arm  maximizing  $\textbf{c}^d(t)^\top \tilde{\mu_k}$ (step 14 in the algorithm), obtain the reward (step 15), and update the parameters of the distributions for each $\tilde{\mu_k}$ (steps 16-21).
 Finally, the reward $r_k(t)$ for choosing an arm $k$ is observed, and relevant parameters are updated.

\subsection{Contextual Bandit with Restricted Context in Non-stationary Environments}
In a stationary environment,  the context vectors and the rewards  are drawn from  fixed probability distributions; the objective is to identify a subset of features allowing for the optimal context-to-arm mapping.
 However, the objective changes when the environment becomes non-stationary.
 
In the {\em non-stationary CBRC setting}, we will assume that {\em the rewards distribution can change only at certain times, and remain stationary between such changes.} Given the non-stationarity of the reward, the player should continue looking for feature subsets which allow for the optimal context-arm mapping, rather than converge to a fixed subset.

Similarly to stationary CBRC problem described earlier, at each iteration, a context $\bf c(t) \in \mathbf{R}^N$ describes the environment,  the player   chooses a subset $C^d \subseteq C$ of  the feature set $C$, and observes the values of those features as a sparse vector  $\mathbf{c^d}(t)$, where all features outside of $C^d$ are set to zero.
Given $\mathbf{c^d}(t)$,  the player chooses an arm $k(t)$. The reward $r_k(t)$ of the played arm is revealed. The reward distribution is non-stationary as described above,  and the (stationary) reward distributions between the change points are unknown. {\em We will make a very specific simplifying assumptions that the change points occur at  every $v$ time points, i.e. that all windows of stationarity have a fixed size.}

\subsubsection{Windows TSRC Algorithm}
\label{subsec:UCB}
Similarly to the TSRC algorithm proposed earlier for the stationary CBRC problem, our algorithm for the non-stationary CBRC uses Thompson Sampling (TS) to find the best $d$ features of the context. 
The two algorithms are practically identical, except for the following important detail: instead of updating the Beta distribution with the number of successes and failures accumulated from the beginning of the game, only the successes and failures within a given stationarity window are counted. The resulting approach is called the {\em Window Thompson Sampling with restricted Context}, or {\em WTSRC}.

Note that $v$, the true size of the stationary window assumed above, is not known by the algorithm, and is replaced by  some approximate window size parameter $w$. In this paper, we  assumed a fixed $w$, and experiment with several values of it; however,   in the future, can be also adjusted using a bandit approach.

\section{Empirical Evaluation}
\label{sec:experimental}
Empirical evaluation of the proposed methods was based on four datasets from the UCI Machine Learning Repository \footnote{https://archive.ics.uci.edu/ml/datasets.html}: Covertype, CNAE-9, Internet Advertisements and Poker Hand (for details of each dataset, see Table \ref{table:Synthetic}).

\begin{table}[t]
	\centering
	\caption{Datasets}
	\resizebox{0.99\columnwidth}{!}{
		\begin{tabular}{ l | c | r | r }
			UCI Datasets                  & Instances    & Features   & Classes \\ \hline
			Covertype                & 581 012      & 95           &  7\\
			CNAE-9                   & 1080       & 857          &  9\\
			Internet Advertisements  & 3279         & 1558         &  2\\
			Poker Hand               & 1 025 010    & 11         &  9\\

		\end{tabular}
	}
	\label{table:Synthetic}
\end{table}

\begin{table*}[tbh]
	\centering
	\caption{Stationary Environment}
	\resizebox{1.84\columnwidth}{!}{
		\begin{tabular}{ l | l | l | l | l | l }
			&  MAB          & Fullfeatures          & TSRC          &   Random-fix   & Random-EI \\ \hline
			Datasets \\ \hline
			Covertype & $70.54\pm 0.30$ &  $ 35.27\pm 0.32$  & $\textbf{53.54}\pm\textbf{1.75}$   & $72.29\pm 2.38$	& $82.69\pm1.87$\\
			CNAE-9 	     &  $79.85 \pm 0.35$    &   $52.01\pm0.20$	  & $\textbf{68.47}\pm \textbf{0.95}$	 & $68.50\pm3.49$	& $80.02\pm 0.23$\\
			Internet Advertisements	      &  $19.22\pm0.20$	                     &   $14.33\pm 0.22$	                & $\textbf{15.21}\pm\textbf{1.20}$	                     & $21.21\pm 1.93$	& $ 23.53\pm\ 1.64$\\
			Poker Hand 	       &  $62.29\pm 0.21$   &   $ 58.57 \pm 0.12$	& $\textbf{58.82}\pm\textbf{0.71}$              & $59.83\pm2.57$   & $58.49\pm0.81$\\
			\hline
		\end{tabular}
	}
	\label{table:AccuSyn}
\end{table*}

\begin{figure*}[tbh]
	\begin{multicols}{2}
		\includegraphics[height=1.4 in,width=0.99\linewidth]{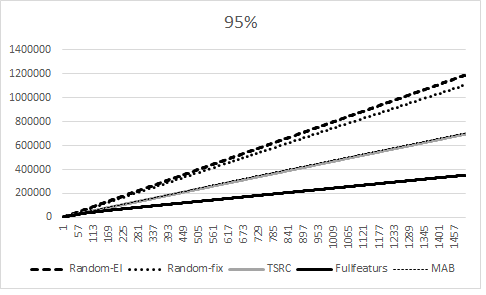}\par
\vspace{0.2in}		
\includegraphics[height=1.4 in,width=0.99\linewidth]{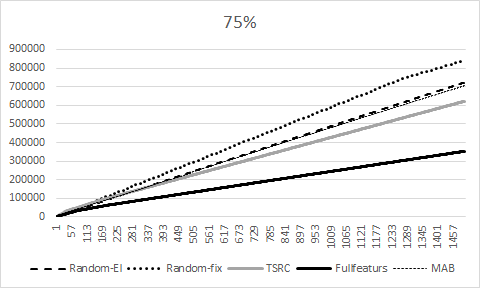}\par
		\includegraphics[height=1.4 in,width=0.99\linewidth]{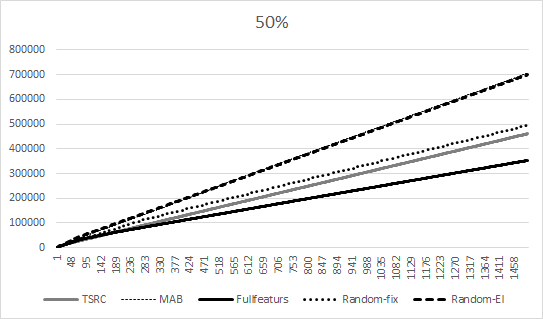}\par
\vspace{0.2in}	
		\includegraphics[height=1.4 in,width=0.99\linewidth]{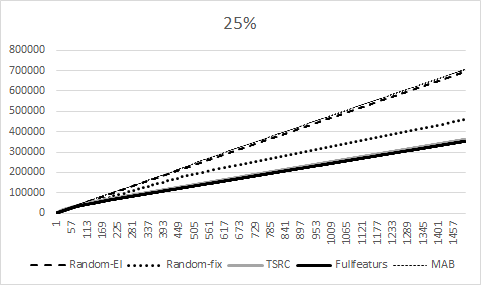}\par
	\end{multicols}
	\caption{Stationary Environment (Covertype dataset)}\label{fig:incrst}
\end{figure*}

\begin{table*}[tbh]
	\centering
	\caption{Non-Stationary Environment}
	\resizebox{1.99\columnwidth}{!}{
		\begin{tabular}{ l | l | l | l | l | l | l }
			&  MAB          & Fullfeatures          & TSRC          &   Random-fix   & Random-EI  & WTSRC\\ \hline
			Datasets \\ \hline
			Covertype  &$69.72\pm 4.30$ &   $ 68.54\pm 2.10$  & $76.12\pm2.86$   & $80.96\pm2.19$	& $ 80.71\pm2.22 $ 	& $\textbf{60.56 }\pm\textbf{2.95}$ \\
			CNAE-9 	 &  $ 74.34 \pm 5.39 $   &  $ 69.21\pm 2.12 $	  & $ 71.87\pm 4.5$	 & $ 79.87 \pm 4.5 $	& $76.21 \pm 1.90$ & $\textbf{65.56 }\pm\textbf{1.05}$  \\
			Internet Advertisements &  $43.99 \pm 3.85$  &  $ 40.21\pm 1.87 $ & $42.01 \pm 1.79 $ & $40.04 \pm4.52 $	& $40.56\pm1.19$ 	& $\textbf{38.06 }\pm\textbf{0.85}$\\
			Poker Hand 	            &  $82.90\pm4.03$ &   $82.44\pm0.43$	 & $78.86 \pm 1.25$ & $79.99 \pm 1.48$ & $79.81 \pm0.48$ 	& $\textbf{77.56 }\pm\textbf{1.79}$ \\
			\hline
		\end{tabular}
	}
	\label{table:AccuSyn1}
\end{table*}

\begin{figure*}[tbh]
	\begin{multicols}{2}
		\includegraphics[height=1.4 in,width=0.99\linewidth]{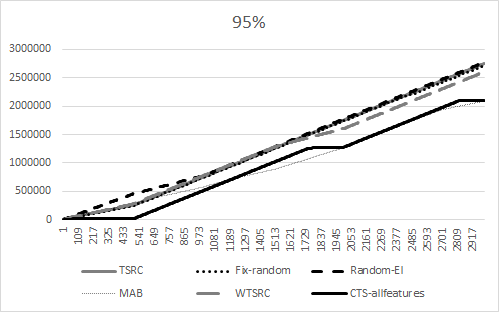}\par
		\vspace{0.2in}
		\includegraphics[height=1.4 in,width=0.99\linewidth]{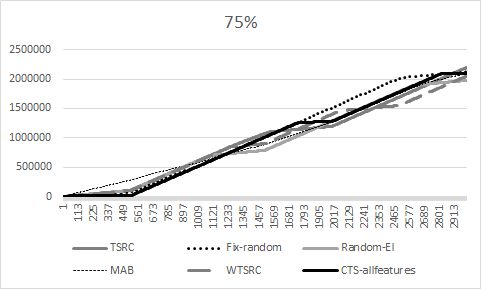}\par
		\includegraphics[height=1.4 in,width=0.99\linewidth]{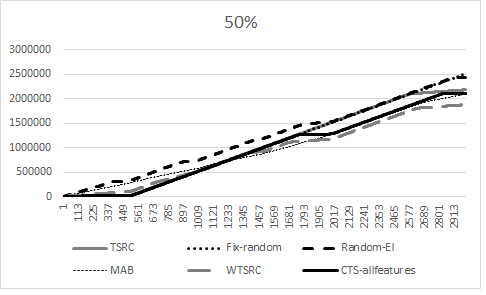}\par
		\vspace{0.2in}
		\includegraphics[height=1.4 in,width=0.99\linewidth]{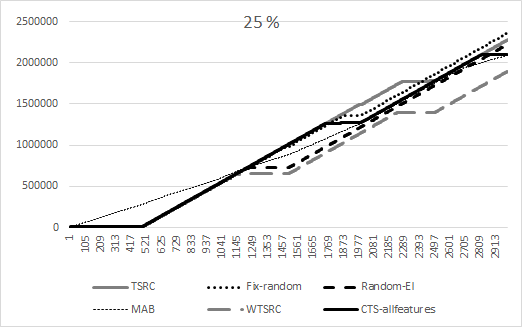}\par
	\end{multicols}
	\caption{Non-Stationary Environment (Covertype dataset)}\label{fig:incr}
\end{figure*}

In order to simulate a data stream, we draw samples from the dataset sequentially, starting from the beginning each time we draw the last sample. At each round, the algorithm receives the reward 1 if the instance is classified correctly, and 0 otherwise.  We compute the total number of classification errors as a performance metric.

We compare our algorithms with the following competing methods:
\begin{itemize}
	\item {\em Multi-arm Bandit (MAB)}: this is the standard Thomspon Sampling approach to (non-contextual) multi-arm bandit setting.
	\item {\em Fullfeatures}: this algorithm uses the contextual Thomspon Sampling (CTS) with the full set of features.
	\item {\em Random-EI}: this algorithm selects a {\em Random} subset of features of the specified size $d$ at {\em Each Iteration} (thus, {\em Random-EI}), and then invokes the contextual bandit algorithm (CTS).
	\item{\em Random-fix}: similarly to {\em Random-EI}, this algorithm invokes  CTS on a random subset of $d$ features; however, this subset is selected once prior to seeing any data samples, and remains fixed.
\end{itemize}

We ran the above algorithms and our proposed TSRC and WTSRC methods, in stationary and non-stationary settings, respectively, for different feature subset sizes, such as 5\%, 25\%, 50\% and 75\% of the total number of features.

In the Figures presented later in this section, we used the parameter, called {\em sparsity}, to denote the percent of features that were {\em not selected}, resulting into the sparsity levels of 95\%, 75\%, 50\% and 25\%, respectively.  In the following sections, we will present our results first for the stationary and then for the non-stationary settings.

\subsection{Stationary case}
Table ~\ref{table:AccuSyn} summarizes our results for the stationary CBRC setting; it represents the average classification error, i.e. the misclassification error, computed as the total number of misclassified samples over the number of iterations. This average errors for each algorithm were computed using 10 cyclical iterations over each dataset, and over the four different sparsity levels mentioned above.

As expected, the CTS algorihtm with the full set of features ({\em Fullfeatures}) achieved the best performance as compared with the rest of the algorithms, underscoring the importance of the amount of context observed in an on-line learning setting.

However, when the context is limited, is in the CBRC problem considered in this work, our TSRC approach shows superior performance (shown in bold in Table ~\ref{table:AccuSyn}) when compared to the rest of the algorithms, except for the {\em Fullfeatures}, confirming the importance of efficient feature selection in the CBRC setting.

Overall, based on their mean error rate, the top three algorithms were TSRC (mean error 49.01\%), Random-fix (mean error 55.46\%), and MAB (mean error 57.98\%), respectively, suggesting that using a fixed randomly selected feature subset may still be a better strategy than not considering any context at all, as in MAB. However, as it turns out, ignoring the context in MAB may still be a better approach than randomly changing the choice of feature at each iteration in {\em Random-EI}; the latter  resulted into  the worst mean error of 61.18\%.

\subsubsection{Detailed analysis on Covertype dataset}
Figure \ref{fig:incrst} provides a more detailed evaluation of the algorithms for different levels of sparsity, on a specific dataset. Ignoring the {\em Fullfeatures} which, as expected, outperforms all methods since it considers the full context, we observe that:

{\bf 95 \% sparsity:} {\em TSRC} has the lowest error out of all methods, followed tightly by {\em MAB}, suggesting that, at a very high sparsity level, ignoring the context ({\em MAB}) is similar to considering only a very small (5\%) fraction of it. Also, as mentioned above, sticking to the same fixed subset of randomly selected features appears to be better than changing the random subset at each iteration.

{\bf 75 \% sparsity:} we observe that {\em Random-EI} has a lower error than the {\em Random-fix}.

{\bf 50 \% sparsity:} {\em TSRC} has the lowest error, followed closely by the {\em Random-fix}, which implies that, at some levels of sparsity, random subset selection can perform almost as good as our optimization strategy. Again, we observe that {\em Random-fix} performs better that the {\em Randon-EI}, where the performance of the latter is close to the multi-arm bandit without any context.

{\bf 25 \% sparsity:} we observe that TSRC perform practically as good as {\em Fullfeatures}, which implies that at this sparsity level, our approach was able to select the optimal feature subset.

\subsection{Non-stationary Case}
In this setting, for each dataset, we run the experiments for 3,000,000 iterations, where we change the label of class at each 500,000 iteration to simulate the non-stationarity. We evaluate our {\em WTSRC} algorithm for the nonstationary CBRC problem against our stationary-setting TSRC method, and against the same five baseline methods we presented before.

Similarly to the stationary case evaluation, the Table ~\ref{table:AccuSyn1} reports the mean error   over all iterations and over the same for level of sparsity as before. Overall, we observed that the {\em WTSRC} performs the best, confirming the effectiveness of using our time-windows approach in a non-stationary on-line learning setting.

Our top three performing algorithms were {\em WTSRC}, {\em Fullfeatures} and {\em TSRC}, in that order, which underscores the importance of the size of the observed context even in the non-stationary on-line learning setting.

\subsubsection{Detailed analysis on Covertype dataset}
Figure \ref{fig:incr} provides a more detailed evaluation of the algorithms for different levels of sparsity, on a specific dataset. The {\em Fullfeatures}, as expected, has the same performance on different level of sparsity, since it has the access to all features. We observe that:

{\bf95 \% sparsity:} {\em MAB} has the lowest error, as compared with the {\em Random-EI} and {\em Random-fix}, which implies that, at high sparsity levels, ignoring the context can be better then even considering a small random subset of it.

{\bf75 \% sparsity:} {\em TSRC} yields the best result, implying that even in a non-stationary environment, a stationary strategy for best subset selection can be beneficial.

{\bf50 \% sparsity:} both {\em Random-Fix} and {\em Randon-EI} yield the worst results, implying that random selection is not a good strategy at this sparsity level in non-stationary environment.

{\bf25 \% sparsity:} our nonstationary method, {\em WTSRC}, outperforms all other algorithms, demonstrating the advantages of a dynamic feature-selection strategy in a non-stationary environment at relatively low sparsity levels.

\section{Conclusions}
\label{sec:Conclusion}
We have introduced a new formulation of MAB, motivated by several real-world applications including visual attention modeling, medical diagnosis and RS.
In this setting, which we refer to as  {\em contextual bandit with restricted context (CBRC)}, a set of features, or a context, is used to describe the current state of world; however, the agent can only choose a limited-size subset of those features to observe, and thus needs to explore the feature space simultaneously with exploring the arm space, in order to find the best feature subset. We proposed two novel algorithms based on Thompson Sampling for solving the CBRC problem in both  stationary and non-stationary environments.
Empirical evaluation on several datasets demonstrates advantages of the proposed approaches. 

\section{Acknowledgments}
The authors thank Dr. Alina Beygelzimer and Dr. Karthikeyan Shanmugam for the critical reading of the draft manuscript.

\renewcommand{\baselinestretch}{1}
\bibliographystyle{named}
\bibliography{ijcai15}

\end{document}